\documentclass[journal,twoclumn]{IEEEtran}
\usepackage{amsmath,amsfonts}
\usepackage{array}
\usepackage[caption=false,font=normalsize,labelfont=sf,textfont=sf]{subfig}
\usepackage{textcomp}
\usepackage{stfloats}
\usepackage{url}

\usepackage{verbatim}
\usepackage{graphicx}
\usepackage{graphicx}%
\usepackage{multirow}%
\usepackage{amsmath,amssymb,amsfonts}%
\usepackage{amsthm}%
\usepackage{mathrsfs}%
\usepackage{xcolor}%
\usepackage{textcomp}%
\usepackage{manyfoot}%
\usepackage{booktabs}%
\usepackage{algorithmicx}%
\usepackage{algpseudocode}%
\usepackage{listings}%
\usepackage{float}

\usepackage[space]{grffile}
\usepackage{latexsym}
\usepackage{longtable}
\usepackage{tabulary}
\usepackage{array}
\usepackage[numbers]{natbib}
\usepackage{url}
\usepackage{hyperref}
\usepackage[utf8]{inputenc}
\usepackage[english]{babel}
\usepackage{bbm}
\usepackage{fancyhdr}
\usepackage[utf8]{inputenc}
\usepackage{setspace}
\usepackage{geometry}
\usepackage{float}
\usepackage{diagbox}
\usepackage{siunitx}
\usepackage{nccmath}

\newcommand{\bd}{\begin{description}}
\newcommand{\ed}{\end{description}}
\newcommand{\ra}{\rightarrow}
\newcommand{\ga}{{\alpha}}

\theoremstyle{plain}%
\newtheorem{theorem}{Theorem}
\newtheorem{proposition}[theorem]{Proposition}%

\theoremstyle{remark}%
\newtheorem{remark}{Remark}%

\theoremstyle{definition}%

\ifCLASSINFOpdf
\else
\fi

\begin{document}
%
\title{Confidence Intervals for\\Evaluation of Data Mining}
%
%
%

\author{Zheng~Yuan
        and~Wenxin~Jiang
\thanks{Zheng Yuan is with Wells Fargo. email: zhengyuan2025@u.northwestern.edu. Wenxin Jiang is with the Department of Statistics and Data Science, Northwestern University, Evanston, IL, 60208, USA. e-mail: wjiang@northwestern.edu.}}

%
%

\markboth{IEEE Transactions on Knowledge and Data Engineering}%
{IEEE Transactions on Knowledge and Data Engineering}
%



\maketitle

\begin{abstract}
In data mining, when binary prediction rules are used to predict a binary outcome, many performance measures are used in a vast array of literature for the purposes of evaluation and comparison. Some examples include classification accuracy, precision, recall, F measures, and Jaccard index. Typically, these performance measures are only approximately estimated from a finite dataset, which may lead to findings that are not statistically significant. In order to properly quantify such statistical uncertainty, it is important to provide confidence intervals associated with these estimated performance measures.
We consider statistical inference about general performance measures used in data mining, with both individual and joint confidence intervals.  These confidence intervals are based on asymptotic normal approximations and can be computed fast, without needs to do bootstrap resampling. We study the finite sample coverage probabilities for these confidence intervals and also propose a `blurring correction' on the variance to improve the finite sample performance. This 'blurring correction' generalizes the plus-four method from binomial proportion to general performance measures used in data mining. Our framework allows multiple performance measures of multiple classification rules to be inferred simultaneously for comparisons.
\end{abstract}


%
\IEEEpeerreviewmaketitle

\section{Introduction}
%
%
%
%
\IEEEPARstart{T}{here} There are many different  performance measures used in data mining, including, for example, accuracy, precision, recall, F measures, Jaccard index. They measure in different perspectives how well a 0/1 valued classification rule predicts a true 0/1 values response. The classification rule can also come from different algorithms, such as nearest neighbor rules, logistic regression, or random forest.  In typical papers, research findings are reported on one or more of the performance measures for one or different classification rules. These performance measures are usually evaluated on a finite validation sample size. Then conclusions are drawn, for example, to compare different rules on how they behave on different performance measures.

For example, the results may be summarized in such a table (Table \ref{tablenoci}). Based on this table, we are tempted to conclude
that the 1-Nearest Neighbor (1NN) rule performs a little worse on the Accuracy measure, and the 
Logistic Regression rule performs much worse on $F_{0.5}$:

\begin{table}[h]
\centering
\caption{Point Estimates of Accuracy and $F_{0.5}$ Measure for Each Classifier based on a Validation Dataset  with 10000 0/1-valued  outcomes}
\begin{tabular}{c*{4}{S}}
  \toprule
  \bfseries  &  \multicolumn{2}{c}{\bfseries Measures}   \\
  \cmidrule(lr){2-3}
  \bfseries Classifiers &  \multicolumn{1}{c}{\bfseries Accuracy} & \multicolumn{1}{c}{\bfseries $F_{0.5}$}  \\
  \midrule
  1-NN & 0.8995 & 0.2477 \\
  Logistic & 0.9354 & 0.0903 \\
  Random Forest & 0.9256  & 0.2738\\
  \bottomrule
\end{tabular}

\label{tablenoci}
\end{table}

There are two statistical research problems related to such a typical practice. One is that the performance on a finite validation dataset carries sample variation. To indicate what the performance of a classification rule will be on the whole population of future datasets, one could attach a margin of error or provide a confidence interval. Secondly, 
the sample variations increase when several sample-based performance measures are used simultaneously to draw multiple conclusions. To account for this variation from such multiple inference, one could consider joint confidence intervals. Joint confidence intervals are known for being valid simultaneously with high probability.

An application of our proposed method (with a special correction to improve finite sample performance) would report these results as follows (see Table \ref{tableci}):

\begin{table}[h]
\centering
\caption{95$\%$ Joint Confidence Intervals (with correction) of Accuracy and $F_{0.5}$ Measure for Each Classifier based on a Validation Dataset  with 10000 0/1-valued  outcomes  }
\begin{small}
\begin{tabular}{c*{4}{S}}
  \toprule
  \bfseries  &  \multicolumn{2}{c}{\bfseries Measures}   \\
  \cmidrule(lr){2-3}
  \bfseries Classifiers &  \multicolumn{1}{c}{\bfseries Accuracy} & \multicolumn{1}{c}{\bfseries $F_{0.5}$}  \\
  \midrule
  1-NN & {(0.8916,~0.9073)} & {(0.2073,~0.2880)} \\
  Logistic & {(0.9289,~0.9418)} & {(0.0322,~0.1485)} \\
  Random Forest & {(0.9187,~0.9324)}  & {(0.2191,~0.3286)}\\
  \bottomrule
\end{tabular}
\end{small}
\label{tableci}
\end{table}

These brackets are  asymptotically 95$\%$ Joint Confidence Intervals, in the limit of large validation data sizes.
We now can examine all six joint confidence intervals together and draw multiple conclusions simultaneously.  For Accuracy, the 1-NN rule performs worse than both joint logistic regression and Random Forest, due to its joint confidence interval being below and non-overlapping with the other two. Similarly, for $F_{0.5}$, the logistic regression performs worse than the other two methods. We are ``asymptotically 95\% confident'' to say that all these findings are statistically valid, whereas before we were not sure any of the observed differences, small or large, are significant statistically.

 By ``asymptotically 95\% confident", we mean that when large enough  data are repeatedly generated from the same unknown probability distribution to construct our joint confidence intervals, these intervals will ALL cover their respective true parameters, and  ANY IMPLIED data dependent conclusions will be correct (simultaneously), at least  94\% times (or anything less than 95\%, since the limiting probability is 95\%).

Below, we give an example probabilistic proof for the above statement, using plain English.  
For example, suppose based on observed data, we  decide to
make two statements jointly: first, among all competing rules, the true  $F_{0.5}$ for Logistic Regression performance is the worst and second, the true Accuracy for 1NN is the worst,  because their respective data-driven  joint confidence intervals lie completely below those of the others rules. 

Then, we will be wrong only when the   true  $F_{0.5}$ for Logistic Regression or the true accuracy for 1NN  is NOT the worst among the competing rules, while their joint confidence intervals lie completely below those of other rules on the respective performance measures. But this would imply that the joint confidence intervals did not cover all the true parameters, which would happen at most about 5\% times.   That is why all our conclusions, making no statement, one statement, or two statements, or more, depending on whatever we may find from the data-based joint confidence intervals, will be correct for  about 95\% times (or more)  in repeated uses of this method.

Unlike in hypothesis testing where one has to be very careful both  in setting up the hypotheses ahead of time and in combining multiple findings, our joint confidence interval approach allows MULTIPLE conclusions to be drawn and these conclusions are drown AFTER looking at these data-driven intervals,  which is a very natural and flexible way to learn from data. Also, when we find that one rule is worse than the other, we could also say how much worse after looking at the gap between the joint confidence intervals, unlike in hypothesis testing where we usually test either for no-difference or a pre-specified difference. 

There are several different ways to make statistical inference on performance measures. For example, F measures have been studied previously with  Bayesian methods (e.g., \cite{bib5}), cross validation (e.g., \cite{bib18}) and bootstrap methods (e.g., \cite{bib7}). Our method provides analytic formulas for the frequentist confidence interval, which is fast computationally, since it does not require resampling. This follows the direction of  \cite{bib8};  \cite{bib4} and \cite{bib14}, who use analytic formulas to compute confidence intervals for the $F_1$ measure. \cite{bib9} considered confidence interval for  the $F_\beta$ measure, \cite{bib19} and \cite{bib11} studied pairwise comparison of classification rules on the $F_1$ measure and the $ F_\beta$ measure, respectively. \cite{bib20} studied pairwise comparison of classification rules on the multi-class $F_1$ measure. 
  Our work allows more general performance measures and also allows simultaneous comparison of more than 2 rules. We also use simulations to study the finite sample coverage probabilities to see if they indeed are close to 95\% and how much wider the joint confidence intervals become  compared   to individual confidence intervals. In addition,
 we find that a correction made on the variance estimate will sometimes be needed for better performance in situations when the asymptotic variance may be nearly singular and it does not make things worse in other easier situations.

Even though we make progresses in considering more general performance measures than the $F_1$ measure and also in considering joint confidence intervals, it is fair to point out that our work is not meant to supersede other recent works on the $F_1$ measure, such as \cite{bib19}, or \cite{bib14}. They considered other aspects that are not covered by this paper. For example, \cite{bib19} considered combining $F_1$ measures on multiple datasets. \cite{bib14} and \cite{bib20} considered multi-class problems with micro- and macro-averaged $F_1$.   It would be interesting future work to study these more sophisticated aspects of their works with other performance measures, using the approach of the current paper. 

\section{Theory and Methodology}
\subsection*{Framework}
We consider the problem of evaluating a   binary decision rule $A$, which is a random variable valued in $\{0,1\}$,  used to predict   a random response variable $Z\in\{0,1\}$. 

We consider inference about a measure of performance of $A$ defined as 

$$M= g(EZA, EA, EZ),$$ where $E$ is the expectation for a random variable. We allow a function $g$ in order to include many examples of performance measures. We provide some examples of $g$ below. In addition, 
we will  also provide formulas for its partial derivatives
$dg=(d_1g,d_2g,,d_3g)$,

which will be
useful later in constructing confidence intervals.  Here we denote $d_r g  = \frac{ \partial g(x_1,x_2,x_3) }{\partial x_r}$ for $r\in\{1,2,3\}$,
where $(x_1,x_2,x_3)=(EZA, EA, EZ)$.

\subsection{Examples of performance measures}

We   can pre-evaluate  $g$ and $dg=(d_1g,  d_2g , d_3g)$   for a list of  performance measures $g$'s that are 
  commonly used in data mining. They are displayed in Table \ref{tablepm} for convenience of the practitioners who want to apply the confidence interval methods in this paper.
  
\begin{table*}[h]
\centering
\begin{small}
\caption{Commonly used performance measures in data mining.}\label{tablepm}
\begin{tabular}{|p{1.5in}|p{1.5in}|p{1.8in}|} 
\hline
Names & Expressions \newline $ g(EZA,EA,EZ)$ & Derivatives  \newline  $ (\frac{\partial g}{\partial EZA}, \frac{\partial g}{\partial EA}, \frac{\partial g}{\partial EZ})$\\
\hline
 Classification accuracy \newline  $ P(A= Z)$ & $2EZA-EA-EZ+1$&$ (2, -1, -1)$\\
 \hline
   Dice similarity or F$_1$ measure &  $ \frac{2EZA}{(E A + E Z)}$ & $ \frac{(1, -0.5 g, -0.5g)}{(0.5EA+0.5EZ)} $\\
\hline
F$_\beta$ measure
(This includes F$_1$ measure as a special case.) & $
\newline\frac{ (1+\beta^2)  EZA}{(EA + \beta^2 EZ)}$ &$\frac{(1, -a g, -bg)}{(aEA+bEZ)} ,$
\newline  where $a=(1+\beta^2)^{-1}$, $b=1-a$.\\
\hline
Jaccard metric \newline
$\frac{P(Z=1,A=1)}
{P(Z=1\ or \ A=1)}$ & $\frac{EZA}{-EZA+  E A +  EZ} $ &  $\frac{(\frac{g(EA+EZ)}{EZA},-g,-g)}{(EA+EZ-EZA) }$ \\
\hline
Tversky  index (\cite{bib15}). \newline (This includes both F$_\beta$ and Jacaard metric as special cases.) &   $\newline\frac{EZA}{((1-a-b)EZA+ a E A + bE  Z)},$ for any $a,b>0$. &  $\newline (\frac{g}{EZA})^2*$ \newline  $(aEA+bEZ,-aEZA,-bEZA)$ \\
\hline
  $\newline$ Correlation \newline $ \frac{cov(Z,A)}{sd(Z)sd( A)}$
 &
 $\newline\frac{EZA-EAEZ}{\sqrt{(EZ -(EZ)^2)(EA -(EA)^2)}}$
 & $\frac{g}{EZA-EAEZ}*$ \newline $(1,\frac{2(EA-1)EZA-EAEZ}{2(EA -(EA)^2)},$\newline
$ 
 \frac{2(EZ-1)EZA-EAEZ}{2(EZ -(EZ)^2)})$\\
 \hline Cosine similarity. (See, e.g., \cite{bib17}.)\newline $ \frac{E(ZA)}{\sqrt{EZ^2}\sqrt{EA^2}}$
 & $ \newline\frac{E(ZA)}{\sqrt{EZ }\sqrt{EA }} $  & $\newline g  * (\frac{1}{EZA},-\frac{1}{2EA},-\frac{1}{2EZ})$  \\
 \hline
  Lift \newline $ P(Z=1|A)/P(Z=1)$ & $E(ZA)/(EAEZ)$ &   $ g * (\frac{1}{EZA},-\frac{1}{EA},-\frac{1}{EZ})$ \\  
  \hline
    Overlap or Szymkiewicz-Simpson coefficient. (See, e.g., \cite{bib17}    2016.)
   & $ \newline EZA/\min(EA,EZ)$ & $ (\frac{1}{min(EA,EZ)}, $\newline $-\mathbbm{1}_{\{EA<EZ\}}\frac{EZA}{(EA)^2},$\newline $-\mathbbm{1}_{\{EZ<EA\}}\frac{EZA}{(EZ)^2})$ (assuming $EA\neq EZ$)
\\
   \hline

\end{tabular}
\end{small}

\end{table*}

To explain why we could include so many performance measures, note that the joint distribution of $(Z,A) \in \{0,1\}^2$ is determined by the four joint probabilities $p(z,a)=P(Z=z,A=a), (z,a)\in\{0,1\}^2$. These four joint probabilities are determined by the three dimensional $(EZA,EZ,EA)$ through $p(1,1)=EZA$, $p(1,0)=EZ-EZA$, $p(0,1)=EA-EZA$, $p(0,0) =1-EZ-EA+EZA$. Therefore, $g(EZA,EZ,EA)$ in fact includes any function of the joint distribution of $(Z,A) \in \{0,1\}^2$. So it covers any performance measure for a binary rule $A$ predicting a binary response $Z$. For constructing confidence intervals, we additionally require $g$ to be differentiable, at the true parameter $(EZA,EZ,EA)$.

\subsection{Asymptotic Normality and Confidence Intervals} 

\begin{proposition}

For $k$ in a finite index set $K$ (such as $K=\{1,2,...,6\}$), let $\hat M_k =  g_k(E_nZA_k, E_nA_k, E_nZ)$ be the estimator of $  M_k =  g_k(E ZA_k, E A_k, E Z)$,   where  $E_n$ is the sample average based on an iid sample $\{((A_k)_i,Z_i) \}_{i=1}^n$ of $(A_k,Z)$ at sample size $n$.
(E.g., $E_nZA_k=n^{-1} \sum_{i=1}^n Z_i(A_k)_i$.)
 
For any $j,k\in K$, let 

$V_{jk} =cov (H_j, H_k)$,

$H_k= d_1g_k  ZA_k + d_2g_k A_k + d_3g_k  Z $ and assume the positive definiteness of the matrix $V=[V_{jk}]_{j,k\in K}$.

For any $r\in \{1,2,3\}$ and $k\in K$, assume the existence of
the partial derivatives

$d_r g_k = \frac{ \partial g_k(x_1,x_2,x_3) }{\partial x_r}$, at
$(x_1,x_2,x_3)= (E ZA_k, E A_k, E Z)$.

Let $\tilde V$ be a consistent estimator of $V$ such that 
$\tilde V_{jk}/  V_{jk} $ converges in probability to 1 for 
all $j,k\in K$, as $n\ra\infty$.

For each $1-\ga\in(0,1)$,  let $q(\ga,R)$  satisfy 

$P[\max_{j\in K} |z_j | < q(\ga,R)]=1-\ga$,

where $(z_j)_{j\in K} \sim N((0)_{j\in K}, R)$ and $R$ is the correlation matrix $(R_{jk})_{j,k\in K}$, defined by $R_{jk}=\frac{  V_{jk} }{ \sqrt{  V_{jj}  V_{kk}} }$ 
for all $j,k\in K$.

\item Use $\Phi$ to denote the standard normal cumulative distribution function.

Then we find that 

 \begin{description}

\item (i) (Asymptotic Normality.) $\sqrt {n} (\hat M_k   - M_k)_{k\in K}   =\sqrt{n}  E_n(H_k-EH_k)_{ k\in K}  +o_p(1) \stackrel{d} \ra   N(0, V) $,
  as $n\ra\infty$.

\item (ii)   (Individual Confidence Intervals.) $\forall \alpha\in(0,1)$, $\lim_{n\ra\infty} P[ M_k \in \hat M_k \pm   \Phi^{-1}(1-\ga/2) n^{-1/2}\sqrt {\tilde V_{kk}}] = 1-\ga$ for  all $k\in K$.
  
\item  (iii)   (Joint Confidence Intervals.) $\forall \ga\in(0,1)$,
let $\tilde q$ be a consistent estimator of $q(\ga,R)$ such that 
$\tilde q$ converges to $q(\ga,R)$ in probability, as $n\ra\infty$, then we have

$\lim_{n\ra\infty} P[ M_k \in \hat M_k \pm \tilde q n^{-1/2}\sqrt {\tilde V_{kk}}\text{ for all $k\in K$}] = 1-\ga$.

\end{description}
\end{proposition}

This is a very general framework that can allow simultaneous inference of possibly different evaluation measures $g_k$ of possibly different classification rules $A_k$.
E.g.,  $A_k=I(S_k>\theta_k)$, where $S_1 = P_1(Z=1|X)$ is estimated by running the 1NN method on a fixed training dataset, evaluated at a future random predictor $X$,   $S_2 = P_2(Z=1|X)$ is estimated by running  logistic regression on a fixed dataset, $S_3 = P_3(Z=1|X)$ is estimated by running  Random Forest on a fixed dataset; $\theta_1=\theta_2=\theta_3=0.5$; $g_1$, $g_2$, $g_3$ are  $F_{0.5}$ scores; 
$(A_4,A_5,A_6)=(A_1,A_2,A_3)$;  $g_4$, $g_5$, $g_6$ are Accuracies.\newline

\begin{remark}
When solving $ q(\ga,R)$ from  

$P[\max_{j\in K} |Z_j | < q(\ga,R)]=1-\ga$, given any  $\ga\in(0,1)$ and any correlation matrix $R$, $(Z_j)_{j\in K} \sim N((0)_{j\in K}, R)$ can be simulated.

One can alternatively use some R package to obtain $q(\ga,R)$, see, e.g.,

\url{https://rdrr.io/cran/mvtnorm/man/qmvnorm.html}

It is also possible to use {\rm invmvnormal "both"} in Stata,
as explained in  \cite{bib6}.\newline

\end{remark}

\begin{remark}
 We will consider two choices for $\tilde V$ and $\tilde q$. Although Choice I is the most straightforward estimation method,  we recommend Choice II, since the numerical studies later suggest that Choice II has better finite sample performances when some asymptotic variances may be close to zero.

 \begin{itemize}
 
 \item Choice I: $\tilde V=\hat V$ and $\tilde q=q(\ga,\hat R)$, where we define, for any $j,k\in K$, and for any $i=1,...,n$,
 
 $\hat R_{jk}=\frac{\hat V_{jk} }{ \sqrt{\hat V_{jj} \hat V_{kk}} }$

 $\hat V_{jk} =(n-1)^{-1} \sum_{i=1}^n   [((\hat H_j)_i - n^{-1}\sum_{i'=1}^n  (\hat H_j)_{i'} )  ( (\hat H_k)_i - n^{-1}\sum_{i'=1}^n (\hat H_k)_{i'} ) ]$,

$ (\hat H_k)_i= (\hat d_1g_k)  Z_i(A_k)_i + (\hat d_2g_k) (A_k)_i + (\hat d_3g_k)  Z_i $,

where for any $r\in \{1,2,3\}$ and $k\in K$,  

$(\hat d_r g_k)= \frac{ \partial g_k(\hat x_1,\hat x_2,\hat x_3) }{\partial x_r}$,

where $(\hat x_1,\hat x_2,\hat x_3)= (E_n ZA_k, E_n A_k, E_n Z)$.

 \item Choice II: 

 $\tilde q=q(\ga,\tilde R)$,
 
  $\tilde R_{jk}=\frac{\tilde V_{jk} }{ \sqrt{\tilde V_{jj} \tilde V_{kk}} }$ for all $j,k\in K$,
  
 $\tilde V=\hat V +D$,
 
 where  $D$ is a diagonal matrix with $D_{kk} = \frac{\Big[\sum_{r=1}^{3}(\hat d_rg_k)^2\Big]\Big(\Phi^{-1}(1-\alpha/2)\Big)^2/2}{n}$, for $k\in K$,  and $\hat V$ is defined as in Choice I.

 \end{itemize}

\end{remark}

\subsection{Proof of Proposition 1}
\begin{description}

\item For (i): This follows from the multivariate  central limit theorem on $(E_nZA_k, E_nA_k, E_nZ)$ (\cite{bib16}, Example 2.18), and
 the delta method on the differentiable function $g_k$'s (\cite{bib16}, Theorems 3.1).
 
\item For (ii): The event in the probability is the same as the event $$|\sqrt{n} (\hat M_k- M_k) /\sqrt{ \tilde V_{kk}}|\leq \Phi^{-1}(1-\alpha/2).$$ Write $\sqrt{n} (\hat M_k- M_k) /\sqrt{ \tilde V_{kk}}=\frac{\sqrt{n} (\hat M_k- M_k) /\sqrt{ V_{kk}}} {\sqrt{\tilde V_{kk} / V_{kk} } }$.
 Results (i) implies that for each $k$, the numerator $\sqrt{n} (\hat M_k- M_k) /\sqrt{ V_{kk}}$ converges in distribution to $N(0,1)$. For any consistent estimator $\tilde V$ of $V$,  the denominator $\sqrt{\tilde V_{kk} / V_{kk}}$ converges to 1 in probability.  By the Slutsky theorem (\cite{bib10}, Theorem 7.15 iii),  their ratio converges in distribution to the ratio of the limits $N(0,1)/1=N(0,1)$. Then $P[ |\sqrt{n} (\hat M_k- M_k) /\sqrt{ \tilde V_{kk}}|\leq \Phi^{-1}(1-\alpha/2)] \ra P[ |N(0,1)|\leq \Phi^{-1}(1-\alpha/2)] = 1-\ga$. 
 \item For (iii): The event of the 
 the left hand side probability can be rewritten as $ \sup_{k\in K} |\frac{\sqrt{n} (\hat M_k- M_k) /\sqrt{ V_{kk}}}{\tilde q \sqrt{\tilde V_{kk} / V_{kk}}}|\leq 1$, where 
 $\tilde q \sqrt{\tilde V_{kk} / V_{kk}}$ converges in probability to
 $q(\ga,R)(>0)$ defined in the statement of the proposition. 
 (Note that $V_{kk}$'s are positive due to positive definiteness of $V$, and the quantile $q(\ga,R)>0$ due to $1-\ga>0$.)

 Note that the left hand side $ \sup_{k\in K} |\frac{\sqrt{n} (\hat M_k- M_k) /\sqrt{ V_{kk}}}{\tilde q \sqrt{\tilde{V}_{kk} / V_{kk}}}|\equiv \sup_{k\in K} |\frac{x_k}{y_k}|$,
 as a mapping of 
 $((x_k)_{k\in K},(y_k)_{k\in K})=(( \sqrt{n} (\hat M_k- M_k) /\sqrt{ V_{kk}} )_{k\in K},
  (\tilde q\sqrt{\tilde{V}_{kk} / V_{kk}})_{k\in K} ))$,  is continuous at any point $((x_k)_{k\in K},(y_k)_{k\in K})$,  if $y_k\neq 0 $ $\forall k\in K$.  Also, $(( \sqrt{n} (\hat M_k- M_k) /\sqrt{ V_{kk}} )_{k\in K},
  (\tilde q\sqrt{\tilde{V}_{kk} / V_{kk}})_{k\in K} ))$  converges in distribution jointly to $(N(0,R), (q(\ga,R))_{k\in K})$ (where $q(\ga,R)$ is a positive quantile), due to 
   a Slutsky theorem similar to \cite{bib10}, Theorem 7.15 (i).
   Then the  
  continuous mapping theorem (\cite{bib10}, Theorem 7.7) implies
  that the $\sup_{k\in K} |\frac{\sqrt{n} (\hat M_k- M_k) /\sqrt{ V_{kk}}}{\tilde q \sqrt{\tilde{V}_{kk} / V_{kk}}}|$ converges in distribution to $\sup_{k\in K} |\frac{N(0,R)_k}{q(\ga,R)} |$.

Now notice that $P[\sup_{k\in K} |\frac{N(0,R)_k}{q(\ga,R)} |\leq 1]=1-\alpha$ due to the definition of $q(\ga,R)$.

Q.E.D.

\end{description}

\subsection{Motivation of the `blurring' correction in Choice II, Remark 2}

The correction in Choice II of Remark 2 does not change the asymptotic behavior since it involves a change of $D$, which is of order $ O_p(1/n)$, applied to an order 1 quantity $\hat V$. However, it will help the finite sample performance if
$\hat V$ is very small in the matrix sense. 

This is motivated by a different way of reckoning the  plus four interval or the Agresti - Coull interval  (see, e.g., \cite{bib1}, \cite{bib2}) 
 as a special case, in order  to make it generalizable to our situation.
When constructing a ($1-\ga$) confidence interval with a good finite sample performance, the Agresti - Coull interval  involves changing the estimator of $V=n var   (E_n Z) $ from the usual   $E_n(1-E_n)$
to $ \tilde E_n(1-\tilde E_n)$, where $\tilde E_n= \frac{nE_n+z^2/2}{(n+z^2 ) } $, $z$ denotes the $(1-\frac{\alpha}{2})$ quantile of normal distribution.   It is called the plus four method when $1-\ga=0.95$, since in that case $z^2\approx 4$.
When $E_n$ or $(1-E_n)$ is small and $n$ is large, the use of $ \tilde E_n(1-\tilde E_n)$ approximately adds
$D=\frac{z^2}{2n}$ to the usual variance estimate $E_n(1-E_n)$ of $V$. This can be alternatively achieved
by injecting a small and independent noise $e_3$ to $E_n Z$, and use its variance  
$\tilde V= nvar  (E_n Z +e_3) = V+D$, where $e_3$ has mean 0 and variance $(\frac{z^2}{2n})/n$. So in this approximation, the Agresti - Coull correction is about the same as considering variance of an error-contaminated estimator
$E_n Z +e_3$ for the parameter $EZ$.

When we generalize this to study $nvar( [\hat M_k]_{k\in K})= nvar(  [g_k(E_nZA_k, E_nA_k, E_nZ)]_{k\in K})$,  for $k\in K$, 
we can have a similar correction of using the variance of a noise-injected estimator of the $M_k$'s: consider $\tilde V=  n var( [g_k(E_nZA_k +e_{1k}, E_nA_k+e_{2k}, E_nZ + e_{3k})]_{k\in K})$, if $\{e_{jk}\}_{j\in\{1,2,3\},k\in K}$ are independent noises with 0 mean and variance $(\frac{z^2}{2n})/n$, it will give $\tilde V\approx V+D$ where $D= diag( [(d_1g_k)^2+(d_2g_k)^2+(d_3g_k)^2] \frac{z^2}{2n} )_{k\in K}$,
by the delta method.

 Due to how this variance correction is related to noise injection, we can call this correction
 method as the ``blurring correction'', which generalizes the  Agresti - Coull method and the plus four method for binomial success probability to inference about many other performance measures.  This ``blurring'' step adds a non-negative diagonal matrix to the variance and `blurs' / increases the width of $O_p(1/\sqrt{n})$-confidence interval by a small amount (of order $1/n$), in order to improve the finite sample performance of the asymptotic normal approximation.

 \section{Simulations}

Our results depend on the large sample asymptotic normal approximation. We will need to verify the finite sample performance by simulation. Some key questions are: can the joint confidence intervals for several parameters be not too much
wider than the individual confidence intervals while doing much better than the individual confidence intervals in terms of the joint coverage probability? Can joint confidence intervals still be useful for typical validation data sizes?

In our simulations, we will consider multiple performance measures, such as the accuracy measure and the $F_{0.5}$ measure,
and compare different classification rules. This is similar to what is done in practice, for example, \cite{bib13}, who estimated both F0.5 scores and Accuracies for different models in his Table 4. 

In the results below, we refer to several kinds of confidence intervals. All are designed to have asymptotic 95\% coverage probabilities:

$\widehat{CI}_k(  S )=\hat M_k\pm q(0.05,[\hat R_{j,k}]_{j,k\in S} )  n^{-1/2}\sqrt{\hat V_{kk}}$, $k\in S$,
use the uncorrected variance estimate $\hat V$   in Choice I of Remark 2, and the index set $S\subseteq K$ is chosen to cover the intended performance parameters $\{M_k,\ k\in S\}$ simultaneously. 

As a special case, $\widehat{CI}_k( \{k\} )=\hat M_k\pm 1.96  n^{-1/2}\sqrt{\hat V_{kk}}$, $k\in K$ are individual confidence intervals.

$\widetilde{CI}_k(  S )=\hat M_k\pm q(0.05,[\tilde R_{j,k}]_{j,k\in S} )  n^{-1/2}\sqrt{\tilde V_{kk}}$, $k\in S$,  where $\hat V$ is replaced by  the   variance estimate $\tilde V$ with the `blurring' correction  in Choice II of Remark 2, and the index set $S\subseteq K$ is chosen to cover the intended performance parameters  $\{M_k,\ k\in S\}$ simultaneously.

\subsection{First simulation model}

Considering a classification problem with two classes, we have $Z\in\{0,1\}$, with probability $P(Z=0)=P(Z=1)=0.5$ and $X|Z=0\sim N(0,1)$, $X|Z=1\sim N(1,1)$. We first sample a data set of 500 $(X,Z)$'s as the fixed training data set. Let $A(x),B(x),C(x)\in\{0,1\}$ be the 1NN rule, logistic regression rule and random forest rule derived from this training data set. For example, $A$ is the 1NN rule derived from the training data set, which means that for each $u$ in the domain of random $X$, $A(u)$ is the $Z$-label of the nearest $X$ observation in the training data set. The $F_{0.5}$ measures of the rules $A$, $B$ and $C$ are $M_1= E(ZA)/(0.2EZ+ 0.8EA)$, $M_3= E(ZB)/(0.2EZ+ 0.8EB)$ and $M_5= E(ZC)/(0.2EZ+ 0.8EC)$, respectively. The Accuracy measures of the rules A, B and C are $M_2 = 1-E(Z-A)^2$, $M_4 = 1 - E(Z-B)^2$ and $M_6 = 1 - E(Z-C)^2$, respectively. In order to simulate true values for the $F_{0.5}$ measures and accuracy measures, we sample 1000000 $(X,Z)$'s, apply the rules $A,~B,~C$ on these samples and evaluate the corresponding performance parameters $M_1,~M_2,...,M_6$. Then we conduct the following experiments to assess the performance of individual and joint confidence intervals based on the methodology in Section II.

\begin{description}
 
\item 1.  The purpose of the first experiment is to check that the individual confidence intervals cover well for individual parameters, but that they do worse and worse when asked to cover more and more parameters simultaneously. We make 10000 simulations. In each simulation, we sample a test data set of size $n=500$ (containing 500 $(X,Z)$ samples) or size $n=2000$. We then apply the 1NN rule, logistic regression rule and random forest rule on each test data set and for each rule, we evaluate its   $F_{0.5}$ measure and Accuracy measure, to estimate the true performance parameters $M_k$, $k=1,...,6$. 
The nominal 95\% individual confidence intervals  $\widehat{CI}_k(\{k\})=\hat M_k\pm 1.96 n^{-1/2}\sqrt{\hat V_{kk}}$, $k=1,...,6$, are provided for each simulated dataset. Then we compute

\begin{itemize}
\item the  coverage probabilities for each parameter individually, i.e., the proportions in 10000 simulations that each parameter is covered by its individual 
confidence interval. This estimates $[P(M_k\in \widehat{CI}_k(\{k\})), k=1,...,6$] to be $[0.944,          0.9433,          0.9454,           0.947,          0.9442,        \\  0.9486 ]$ for $n=500$ and $[0.9471,          0.9513,          0.9494,          0.9482,   \\       0.9474,          0.9508]$ for $n=2000$.  We see that these individual CIs have coverage probabilities quite close to the nominal 95\% even for a relatively small sample size $n$.

\item We compute some examples of the  coverage probabilities for two parameters simultaneously, which estimates  [$P(M_k\in \widehat{CI}_k(\{k\}),  k\in\{1,2\})$, $P(M_k\in \widehat{CI}_k(\{k\}),  k\in\{3,4\})$, $P(M_k\in \widehat{CI}_k(\{k\}),  k\in\{5,6\})$] to be [0.9125 , 0.9162, 0.9166], for $n=500$, and [0.9205, 0.9204, 0.9203] for $n=2000$.    We see that  these individual CIs have joint coverage probabilities lower than the nominal 95\% even for a relatively large sample size $n$.

\item We compute  the  coverage probabilities for 6 parameters simultaneously, which estimates 
$P(M_k\in \widehat{CI}_k(\{k\}), \  k=1,...,6)$ to be 0.8495  for $n=500$, and 0.8572 for $n=2000$.
We see that  these individual CIs now have a much lower joint coverage probabilities,  when asked to cover 6 parameters simultaneously.

\end{itemize}
   
 \item 2. The previous experiment convinces us about the need to use joint confidence intervals whenever we want to cover multiple parameters simultaneously.   We now consider joint CIs covering two parameters simultaneously. For each of the 1-NN rule, logistic regression rule and random forest rule, we evaluate its 95\% joint confidence intervals for the $F_{0.5}$ measure and the Accuracy measure. We also evaluate the average interval lengths of two 95\% joint confidence intervals. Then for each rule, we calculate the joint coverage probabilities (the proportions in 10000 simulations that the  joint confidence intervals for $F_{0.5}$ measure and accuracy measure cover the true values simultaneously). We also calculate the average of average interval lengths from 10000 simulations.

 \item 3. We also consider the 6 performance parameters (for 3 rules $\times$ 2 measure) together in the sense that we evaluate the 95\% joint confidence intervals for these 6 measures. In each simulation, we evaluate the average interval lengths of these six 95\% simultaneous confidence intervals. Then we compute the simultaneous coverage probabilities and the average (with respect to 10000 simulations) of average (with respect to 6 performance parameters) interval lengths of joint confidence intervals.

The numerical results are shown in Table \ref{tablesimu500} and Table \ref{tablesimu2000}. We can see that in both the $n=500$ test sample size case and the $n=2000$ test sample size case, our 95\% joint confidence intervals for two measures of each rule as well as for all the six performance parameters (for 3 Methods $\times$ 2 Measures) have nearly 0.95 coverage proportions. The  interval lengths in Table \ref{tablesimu2000} are shorter than the interval lengths in Table \ref{tablesimu500}, due to increased sample size $n$. 

\end{description}

\begin{table*}[h]
\caption{Coverage Probabilities and Average Lengths for Joint Confidence Intervals based on Simulated Data (Test Sample Size=500)}
\begin{small}
\begin{tabular}{||c|c|c|c|c||} 
 \hline
 Classifier & True $F_{0.5}$ Measure & True Accuracy &Joint Coverage & Average Length \\ [0.5ex] 
 \hline
  1-NN  & 0.6056 & 0.6060 & 0.9460 & 0.1075  \\
 \hline
 Logistic Regression & 0.6905 & 0.6909 & 0.9468 & 0.0998 \\
  \hline
Random Forest & 0.6057 & 0.6064 & 0.9447 & 0.1075  \\
 \hline
3 Methods $\times$ 2 Measures& \slashbox{}{} & \slashbox{}{} & 0.9453 & 0.1168 \\
\hline
\end{tabular}
\label{tablesimu500}
\end{small}
\end{table*}

\begin{table*}[h]
\caption{Coverage Probabilities and Average Lengths for Joint Confidence Intervals based on Simulated Data (Test Sample Size=2000)}
\begin{small}
\begin{tabular}{||c|c|c|c|c||} 
 \hline
 Classifier & True $F_{0.5}$ Measure & True Accuracy &Joint Coverage & Average Length \\ [0.5ex] 
 \hline
 1-NN & 0.6056 & 0.6060 & 0.9490 & 0.0538  \\
 \hline
 Logistic Regression & 0.6905 & 0.6909 & 0.9481 & 0.0499 \\
  \hline
Random Forest & 0.6057 & 0.6064 & 0.9492 & 0.0538  \\
 \hline
3 Methods $\times$ 2 Measures& \slashbox{}{} & \slashbox{}{} & 0.9460 & 0.0584 \\
\hline

\end{tabular} 
\end{small}
\label{tablesimu2000}
\end{table*}

\subsection{Second simulation model}
The second simulation model is using the empirical distribution of a real data set as the underlying true data generating model and draws  bootstrap samples from it (with replacements) as the simulated data sets. Such a simulation model based on resampling a real dataset may be more realistic and may give ``surprises'' that the researchers were unprepared for. In our case, we found that the finite sample performance of the asymptotic confidence intervals are
not so good, probably due to the asymptotic variance being close to singular,  so a correction method is proposed and it can lead to significant improvement.

A data set from UCI database (\cite{bib3}) for predicting the number of rings  (related to the ages) of abalones (a group of sea snails), is composed of 4,177 samples described by 9 features (\cite{bib12}). To illustrate our method, we consider a two-class classification problem (6 rings vs. other number of rings, following \cite{bib19}). We first randomly select a train set of size 844 from the original dataset, and train 1-NN, Logistic Regression and Random Forest algorithm, separately on the train set and obtain three prediction rules.  
We then compare how these three different rules perform according to the $F_{0.5}$ measure and the Accuracy measure, when applied to a data generating process defined by the  empirical distribution of the remaining $3333$ data points.
 \footnote{ We saved a larger proportion of data for this purpose  only because in this way the empirical distribution will approximate the underlying data generating process more realistically. Otherwise there is no consequence to this choice, since we can sample test data with $any$ sample size $n$  from this empirical distribution. We also did check that the training data size $844$ is big enough, so that increasing it does not alter the results qualitatively.} 
   We treat the $F_{0.5}$ measures and Accuracy measures for the three classification rules when applied to this empirical distribution as the ``true values''. This will give 6 true performance parameter values 
$M_1,...,M_6$, arranged in the same way as in the first simulation model.

Next, we sample, with replacement, a test set of size $n=3333$ from the original test set, and compute various kinds of confidence intervals. Repeating this resampling scheme 10000 times. Then we can compute the coverage probabilities
of these confidence intervals and their average lengths.

\begin{description}
 \item 1. The purpose of the first experiment is to check that the individual confidence intervals cover well for individual parameters. Consider
95\% individual confidence intervals  $ \widehat{CI}_k(\{k\})=\hat M_k\pm 1.96 n^{-1/2}\sqrt{\hat V_{kk}}$, $k=1,...,6$. Then we compute
  the  coverage probabilities for each parameter individually, i.e., the proportions in 10000 simulations that each parameter is covered by its individual confidence interval. This estimates $[P(M_k\in \widehat{CI}_k(\{k\})), k=1,...,6$] to be
$[0.9442,           0.948,          0.9006,           0.948,          0.9262 ,   \\      0.9472]$ for $n=3333$.  

  We see that the   individual CI for logistic regression rule on $F_{0.5}$ measure ($M_3$) 

does not have coverage probability being very close to the nominal 95\%, even for relatively large sample size $n=3333$.  

This may be because the asymptotic variance sometimes becomes close to
zero.
\\
\item 2. In the second experiment, we consider joint confidence intervals 
$\widehat{CI}_k( \{3,4\} )=\hat M_k\pm q(0.05,[\hat R_{j,k}]_{j,k\in\{3,4\}} )  n^{-1/2}\sqrt{\hat V_{kk}}, ~~k\in \{3,4\},$ for jointly estimating
the two performances measures $M_{3,4}$ for the logistic regression rule. Based on 10000 simulations, we estimate the coverage of the nominal 95\% joint confidence intervals $P[M_3\in \widehat{CI}_3( \{3,4\} ), \  M_4\in \widehat{CI}_4( \{3,4\})]$ to be only 0.9066, for $n=3333$.   This may be because the asymptotic variance sometimes becomes close to  singular. 
[The individual confidence interval $\widehat{CI}_k(\{k\})$'s perform even worse:
   $P(M_k\in \widehat{CI}_k(\{k\}) , k=3,4)$ is estimated to be only 0.8586, for $n=3333$.]
\\
\item 3. In the third experiment, we consider joint confidence intervals 
$\widehat{CI}_k( \{1,...,6\}  )=\hat M_k\pm q(0.05,[\hat R_{j,k}]_{j,k\in\{1,...,6\}} )  n^{-1/2}\sqrt{\hat V_{kk}},~~k\in \{1,...,6\}, $for jointly estimating $M_{1,...,6}$, including 
both performances measures  for all three rules.

Based on 10000 simulations, we estimate   
$P[M_k\in \widehat{CI}_k (\{1,...,6\} ) , k=1,...,6]$  to be only 0.8752, for $n=3333$, 
for the nominal 95\% joint confidence interval.  
 This may be because the asymptotic variance sometimes becomes close to  singular. 
 [The individual confidence interval $\widehat{CI}_k(\{k\})$'s perform even worse:
   $P(M_k\in \widehat{CI}_k(\{k\}) , k=1,...,6)$ is estimated to be only 0.7193, for $n=3333$.]
\end{description}

When we use the corrected confidence intervals  
$\widetilde{CI}_k(  S )=\hat M_k\pm q(0.05,[\tilde R_{j,k}]_{j,k\in S} )  n^{-1/2}\sqrt{\tilde V_{kk}}$, $k\in S$,  where $\tilde V$ follows Choice II of Remark 2, and the index set $S$ is chosen to cover the intended parameters simultaneously, the coverage probabilities are no longer so much lower than the nominal level 0.95.  The corrected individual confidence intervals now have coverage 
 $[P(M_k\in \widetilde{CI}_k(\{k\})), k=1,...,6$] estimated to be $[0.9496,          0.9546,          0.9726,    0.9552,          0.9398,          0.9518]$ for $n=3333$. 

The results for the corrected joint confidence intervals are summarized in Table \ref{tableabalone}, where we also check whether using the joint confidence intervals with or without the correction leads to too much inflation in the widths of the confidence intervals.  We can see that our 95\% joint confidence intervals for two measures of each rule ($S=\{1,2\}$, $S=\{3,4\}$, $S=\{5,6\}$), as well as for all the six performance parameters (for 3 Methods $\times$ 2 Measures, $S=\{1,...,6\}$), now all have good coverage probabilities after the correction, while still keeping relatively decent average interval lengths. 

\begin{table*}[h]
\caption{Joint Coverages and Average Lengths for Joint and Individual Confidence Intervals based on Abalone Data}

\begin{small}
\begin{tabular}{||c|c|c|c|c|c||} 
 \hline
 Classifier & True $F_{0.5}$   & True Accuracy  & Method & Joint Coverage & Average Length\\  
 \hline
   &   &   & corrected &0.9558 & 0.0877 \\ 
  $1NN$ & 0.2537  & 0.8992 &joint & 0.9437   & 0.0848  \\  
 & & & individual & 0.8963&0.0748\\
  
 \hline
   & &  & corrected &  0.9600 & 0.0948  \\ 
  Logistic Regression & 0.0654 & 0.9331&joint & 0.9066  & 0.0879   \\ 
 & & & individual &0.8586&0.0771\\
 \hline
    &  &   & corrected & 0.9384 & 0.1089 \\ 
   Random Forest & 0.2745 &  0.9253&joint & 0.9315   & 0.1074  \\  
 &  & & individual &0.8768&0.0946\\
 \hline
  & & & corrected &  0.9472 & 0.1014  \\ 
 3 Methods $\times$ 2 Measures & \slashbox{}{} & \slashbox{}{}&joint & 0.8752   & 0.0917  \\   
 &  & & individual &0.7193&0.0694 \\
 \hline
\end{tabular}
\end{small}

\label{tableabalone}
\end{table*}

\subsection{Third simulation model}

We now double the number of performance measures to see if the proposed method is still effective and reliable. We consider a total of 12 performance parameters (for 4 rules × 3 measures). The 4 classification rules are: 1NN, Logistic regression, Random Forest, and Support Vector Machine (SVM). For each classification rule, we consider 3 commonly used evaluation measures: $F_{0.5}$, Accuracy and Lift.  It is not uncommon to study so many performance parameters together in the literature. For example, \cite{bib19} compared classification rules from 4 different algorithms using the $F_1$ measure. \cite{bib14} 's Table 4 includes 4 classification rules and 2 variations of the $F_1$ measure.

In this simulation, we also use a new data generating process from a different real data set to see if the proposed method still works. The dataset is the well-known letter recognition dataset available at the University of California Irvine (UCI) Machine Learning Repository (\cite{bib3}). Certain extraction and distortion techniques are implemented for each letter to produce 16-attribute values in a dataset with 20000 instances. We now consider a binary classification problem. Following \cite{bib19}, we combine letters A and B together in a group and label it to be the positive class, while other letters are combined in another group, labeled as the negative class.

Similar as the scheme in the second simulation model in Section 3.2, we first randomly select a subset of size 3936 from the original dataset and train the 1NN, Logistic Regression, Random Forest and SVM algorithms, separately on this subset, to obtain four prediction rules.  The empirical distribution of the rest 16064 data will be used as the true distribution $P$ of the data generating process in our simulations. 
We then compute the $F_{0.5}$ measures, Accuracy measures and Lift measures according 
to the distribution $P$ for the four prediction rules. This will give a total of $4\times 3=12$ true performance parameter values, denoted as $M_1,...,M_{12}$. 

Next, we sample (with replacement) a test set of size $n=3000$ from the distribution $P$, and compute various kinds of confidence intervals. Repeating this sampling scheme 10000 times. Then we can compute the coverage probabilities
of these confidence intervals and investigate their lengths. We conduct the following experiments:

\begin{description}
\item 1. In the first experiment, we consider joint confidence intervals for simultaneously estimating the three performances measures $M_{4,5,6}$ for the logistic regression rule:
$
\widehat{CI}_k( \{4,5,6\} )=\hat M_k
\pm q(0.05,[\hat R_{jk}]_{j,k\in\{4,5,6\}} )  n^{-1/2}\sqrt{\hat V_{kk}},~~k\in \{4,5,6\}. 
$
Based on 10000 simulations, we estimate the coverage of the nominal 95\% joint confidence intervals $P[M_4\in \widehat{CI}_4( \{4,5,6\} ), ~M_5\in \widehat{CI}_5( \{4,5,6\} ), ~M_6\in \widehat{CI}_6( \{4,5,6\} )]$ to be only 0.9155, for $n=3000$. 

[The individual confidence interval $\widehat{CI}_k(\{k\})$'s perform even worse: $P(M_k\in \widehat{CI}_k(\{k\}) , k=4,5,6)$ is estimated to be only 0.8643 for $n=3000$.] When we use the corrected confidence intervals: 
$\widetilde{CI}_k(  S )=\hat M_k\pm q(0.05,[\tilde R_{jk}]_{j,k\in S} )  n^{-1/2}\sqrt{\tilde V_{kk}},~~k\in S,$  where $\tilde V$ follows Choice II of Remark 2 and the index set $S$ is chosen to cover the intended parameters simultaneously, the coverage probability becomes 0.9410 and it is now closer to the nominal level 0.95. For other classification rules, where the coverages of the nominal 95\% joint confidence intervals have already been close to 0.95 (0.9485 for $1NN$, 0.9430 for Random Forest and 0.9445 for SVM), employing the corrected confidence intervals does not affect the coverages much.
\\
\item 2. In the second experiment, we consider joint confidence intervals for simultaneously estimating $M_{1,...,12}$, that is, for all three performances measures  for all four rules:
$\widehat{CI}_k( \{1,...,12\}  )=\hat M_k\pm q(0.05,[\hat R_{jk}]_{j,k\in\{1,...,12\}} )  n^{-1/2}\sqrt{\hat V_{kk}},~~\\ k\in \{1,...,12\}.$

Based on 10000 simulations, we estimate the coverage of the nominal 95\% joint confidence intervals 
$P[M_k\in \widehat{CI}_k (\{1,...,12\} ) , k=1,...,12]$  to be 0.9290, for $n=3000$. 
By comparison, the individual confidence interval $\widehat{CI}_k(\{k\})$'s perform worse: $P(M_k\in \widehat{CI}_k(\{k\}) , k=1,...,12)$ is estimated to be only 0.7370, for $n=3000$. Again, when we use the corrected confidence intervals, the coverage probability becomes 0.9513 and it is closer to the nominal level 0.95. In this case, we have doubled the total number of performance parameters compared with Section 3.2, and the results from the experiment indicates that the proposed method is still reliable.
   
\end{description}

The numerical results for the different kinds of confidence intervals are summarized in Table \ref{tableletter}, where we can also see that using the joint confidence intervals with or without the correction does not lead to too much inflation in the lengths of the confidence intervals compared to the individual ones.
Here, we did not directly list the average interval lengths themselves, since the Lift values are much larger than the other two evaluation measures.
Instead, we list the average rescaled lengths, where each length is divided by the corresponding true parameter value.

\begin{table*}[h]
\caption{Joint Coverages and Average Rescaled Lengths for Joint and Individual Confidence Intervals based on Letter Data}

\begin{small}
\begin{tabular}{||c|c|c|c|c|c|c||} 
 \hline
 Classifier & True $F_{0.5}$   & True Accuracy  & True Lift  & Method & Joint Coverage & Average(Length/True)\\  
 \hline
 &    &    & & corrected & 0.9597 &  0.1311 \\ 
 $1NN$& 0.8909 & 0.9838 & 11.5482 &joint &  0.9485   &  0.1253  \\  
 & &  && individual & 0.8950 & 0.1093 \\
  
 \hline
 &   &    & & corrected & 0.9410   & 0.2081   \\ 
  Logistic Regression&0.5853 & 0.9384 &8.7963&joint &  0.9115  &  0.1970  \\ 
 & &  && individual & 0.8643& 0.1781 \\
 \hline
    &    &    & & corrected & 0.9617  & 0.1285  \\ 
  Random Forest &0.9319 & 0.9803 &12.9096&joint & 0.9430    &   0.1176 \\  
 &  &  & & individual & 0.8980& 0.1068\\
 \hline
    &   &     && corrected & 0.9553  & 0.1375  \\ 
  SVM &0.8871 & 0.9704 &12.9563&joint &  0.9445   &  0.1263 \\  
 &  &  && individual &0.9057 & 0.1148 \\
\hline
  &  & & & corrected & 0.9513  & 0.1794 \\ 
  4 Methods $\times$ 3 Measures & \slashbox{}{}& \slashbox{}{} &\slashbox{}{} &joint & 0.9290   & 0.1670  \\   
 &  &  & & individual &0.7370& 0.1276\\
 \hline
\end{tabular}
\end{small}
\label{tableletter}
\end{table*}

\section{Discussions}
The proposed method involves confidence intervals that have analytic formulas that can be easy to compute. Joint confidence intervals allow flexible inference after seeing the data and allow multiple number of conclusions to be drawn with a joint `level of confidence'. In the era of big data, large-sample scenarios are not rare. With large enough sample sizes, the joint confidence intervals can still lead to detection of statistically significant differences. The correction method we propose generalizes the plus four method and improves finite sample performance when the asymptotic variance may be close to zero in the matrix sense. 

Even though our method emphasizes on joint confidence intervals, our results on individual confidence intervals are also relatively new and not so commonly applied yet. Our method can be applied to most performance measures used in data mining. We just need to put the performance measure  into a functional form $g$, and evaluate its partial derivatives. 
Due to multiple virtues of this work, we expect that this work can be widely applied in the practice of data mining.

There are several possibilities for improvement or generalization of the current work. Often, there is a natural range for the performance measure. E.g., the F measures are valued in [0,1]. It may be possible to improve the confidence interval either by truncating it to [0,1], or by applying a transformation to $\Re$ (e.g., by a quantile transform), to improve the coverage. The current work only covers binary classification. It will be interesting future work to investigate how to find confidence intervals for general performance measures in the multi-class or regression settings.  For example, one may consider extending \cite{bib14}'s work from $F_1$ measure to other measures,  treating multi-class problems with micro-averaging and macro-averaging.

It may also be possible to apply high dimensional central limit theorems, such as resampling methods that allow the number of simultaneous confidence intervals to grow exponentially with respect to sample size (see e.g., \cite{bibhidim} ). Currently,  however, we only focus on analytic formulas derived from classical asymptotic theory. This is a very natural practice in the area of evaluation of data mining rules from a few competing methods (see, e.g., \cite{bib14}), since the number of confidence intervals is typically low compared to the sample size,  and resampling methods would require much more computation. 

There is another possible extension of the current work. A classification rule, say, $A$,  is typically learned from applying a method (e.g. logistic regression) to a set of training data. Currently, our work applies to finding how this particular classification rule $A$ will perform on future data, but not on how the method behind it, such the logistic regression, behaves, when applied to many different training datasets and obtaining many different classification rules. In a simple way, we say that we only study the performance of the `rule', not of the `method'. It is much more difficult and problem-specific to do the latter. In the logistic regression case, there may be a way to use the delta method to incorporate the variation from the training data to the confidence intervals for the performance of the method. However, there are other cases when such a delta method will not work, since we do not know whether asymptotic normality can be applied to the training step. For example, some methods are non-parametric (such as the 1NN) and some parametric methods (such as deep neural net) do not necessarily converge to the extent of being able to apply the asymptotic normality. We leave these as possible future works. A preliminary attempt using subsampling is reported in \cite{bib22}.

\section*{Acknowledgements}
This paper is partially based on the Ph.D. Dissertation \cite{bib22} of the first author. We thank Dr. Wei Ju for providing some related R programs. We also thank Professor Hongmei Jiang and Professor Noelle Samia for helpful discussions. We are also grateful for the encouragements and helpful discussions from the late Professor Martin Tanner.

\section*{Declarations}

\subsection*{Conflict of interest/Competing interests}

All authors certify that they have no affiliations with or involvement in any organization or entity with any financial interest or non-financial interest in the subject matter or materials discussed in this manuscript.
The authors have no competing interests to declare that are relevant to the content of this article.

\subsection*{Funding} 
The authors did not receive support from any organization for the submitted work.

\subsection*{Author contributions}
Each listed author has made a significant contribution to the research presented in the submitted manuscript. Methodology development and mathematical derivation were primarily conducted by Wenxin Jiang. All the simulations and data analyses were primarily performed by Zheng Yuan. Both of the authors contributed on writing the manuscript. Both of the authors have read and approved the contents of this paper.

\subsection*{Data availability}
The  `Abalone' data and `Letter Recognition' data that support the findings of this study are available in the  the University of California Irvine (UCI) Machine Learning Repository (\cite{bib3}) with the identifier \url{https://doi.org/10.24432/C55C7W} and \url{https://doi.org/10.24432/C5ZP40}, respectively. All other relevant data generated and analysed during this study, which include experimental and computational data, are included in this article.


\ifCLASSOPTIONcaptionsoff
  \newpage
\fi



\bibliography{bibliography}
\bibliographystyle{IEEEtran}
\end{document}